\documentclass[11pt]{article}
\usepackage[preprint]{acl}

\usepackage{times}
\usepackage{latexsym}
\usepackage[T1]{fontenc}
\usepackage[utf8]{inputenc}
\usepackage{microtype}
\usepackage{inconsolata}

\usepackage{amsmath,amssymb}
\usepackage{graphicx}
\usepackage{booktabs}
\usepackage{tabularx}
\usepackage{multirow}
\usepackage{xcolor}
\usepackage{colortbl}
\usepackage{pifont}
\usepackage{placeins}
\usepackage{wrapfig}

\newcommand{\best}[1]{\textbf{#1}}

\title{EASE-KV: Diagnosing the Gap Between Memory Availability and Effective Use}

\author{
    \textbf{Di Lu}\textsuperscript{1} \quad
    \textbf{Zechen Li}\textsuperscript{1} \quad
    \textbf{Xuanyu Yue}\textsuperscript{1} \\
    \textbf{Jiabo Fan}\textsuperscript{1} \quad
    \textbf{Hao Xue}\textsuperscript{2} \quad
    \textbf{Flora D. Salim}\textsuperscript{1} \\[0.6em]
    \textsuperscript{1}University of New South Wales, Sydney, Australia \\
    \textsuperscript{2}Hong Kong University of Science and Technology (Guangzhou), China \\[0.6em]
    \texttt{di.lu1@student.unsw.edu.au} \quad
    \texttt{zechen.li@unsw.edu.au} \\
    \texttt{xuanyu.yue@student.unsw.edu.au} \quad
    \texttt{z5577264@zmail.unsw.edu.au} \\
    \texttt{haoxue@hkust-gz.edu.cn} \quad
    \texttt{flora.salim@unsw.edu.au}
}

\begin{document}

\maketitle
\raggedbottom

% Abstract
\begin{abstract}
Reusable key--value (KV) memory avoids repeatedly prefilling retrieved text, but stored and injected memories do not necessarily become usable evidence. We present EASE-KV, a controlled framework for understanding this gap under fixed candidates and a frozen language model. EASE-KV separates three interfaces: whether required relations are preserved during encoding, whether preserved memories receive appropriate influence during allocation, and whether outcome supervision can identify useful allocations. Matched experiments show that causal encoding improves compositional accuracy by 25.0 percentage points over block-diagonal encoding, while the tested inference-time bridging methods fail to recover the lost contextual information. With unchanged stored states, diagnostic evidence-guided allocation improves LongMemEval accuracy from 31.2\% to 38.4\%, demonstrating that memory influence can be controlled without re-encoding. However, controllability does not guarantee learnability: binary correctness provides informative reward contrast for 53.0\% of sampled query groups, compared with 73.6\% under the evaluated shaped-reward setting. These findings show that effective KV memory requires more than retrieval and injection: the required relations must be encoded, preserved memories must receive appropriate influence, and supervision must distinguish better allocations from worse ones.
\end{abstract}

% Main text
\section{Introduction}
\label{sec:intro}

Long-term assistants must reuse prior information without repeatedly placing
an ever-growing history in the prompt. Text retrieval ranks and serializes
memories for each query~\cite{rag,li2026zara}, whereas reusable key--value (KV) memory
encodes them once and injects their attention states at inference, reducing
repeated prefill~\cite{promptcache,cacheblend,blockattention,kvlink,inferscale}.
Yet a retrieved memory can be present in the cache without correctly
influencing the answer. We call this gap between memory availability and
effective use \emph{reuse without recall}.

This failure arises because latent reuse freezes representations before they
are needed. Unlike text retrieval, which exposes selected memories to
contextual prefill, KV injection combines states encoded under potentially
different contexts. Their relative influence depends on keys, positions, and
the current query. Reuse therefore separates when a relation is formed from
when it must be used.

Existing work addresses parts of this pipeline but does not isolate their
effects. KV-reuse systems improve cache composition through recomputation,
blockwise attention, link tokens, or contextual encoding
~\cite{promptcache,cacheblend,blockattention,kvlink,inferscale}, yet end-to-end
accuracy cannot establish whether the required relation was encoded.
Attention steering amplifies selected context without updating the language
model~\cite{pasta,autopasta,spotlight}, but assumes the needed information is
already represented. Memory agents jointly optimize storage, retrieval, and
answering from task outcomes~\cite{memoryr1}, obscuring whether failures arise
from candidate selection, memory use, or insufficient reward contrast.

These failure sources require different remedies: allocation cannot recover
a relation missing from the cache, while expressive control is ineffective
if feedback cannot distinguish better settings. We introduce
\textbf{EASE-KV}, an \textbf{E}ncoding--\textbf{A}llocation--
\textbf{S}upervision \textbf{E}valuation framework that separates these
factors under matched candidates and a frozen language model. Cross-memory
visibility tests what encoding preserves; signed attention-logit biases
control each memory's influence without changing its stored KV states; and
within-query reward contrast measures whether outcome feedback can teach
this control. EASE-KV asks three questions in sequence:
\emph{Is the information represented? Can its use be controlled? Can the
control be learned?}

This decomposition identifies where system improvements are needed. Better
allocation cannot compensate for impoverished representations, repeated
encoding sacrifices reuse efficiency, and greater policy capacity cannot
replace missing learning signals.

Our contributions follow this sequence:
\begin{enumerate}
    \item \textbf{Encoding fidelity.} Matched-visibility experiments show that
    causal encoding improves compositional accuracy by $25.0$ points over
    block-diagonal encoding. Broader encoding context improves relational
    fidelity, whereas the tested inference-time bridges fail to recover
    the contextual information lost in independently cached states.

    \item \textbf{Allocation headroom.} Signed per-memory biases separate
    representation from use. Diagnostic evidence-guided allocation improves
    LongMemEval accuracy by $7.2$ percentage points without changing
    candidates or stored KV states. In temporal conflict, hand-specified
    recency recovers $19.5$ of the $20.9$-point text--latent gap, showing
    that allocation can select between preserved versions when the
    appropriate preference is specified.

    \item \textbf{Supervision resolution.} Binary correctness provides
    within-query reward contrast for $53.0\%$ of sampled groups, compared
    with $73.6\%$ under the evaluated shaped-reward setting. Learned
    allocation policies show limited improvements over a strong
    evidence-guided reference, highlighting that the ability to control
    memory influence does not guarantee a transferable allocation policy.
\end{enumerate}

Together, these results establish that encoding, allocation, and supervision
are complementary but non-substitutable. Reusable KV memory requires the
relevant relation to be represented, appropriately weighted, and supported
by feedback that distinguishes better use. This perspective complements
faster cache serving by identifying the semantic and learning conditions
needed for latent state to function as long-term memory.
\section{Related Work}
\label{sec:related}

\paragraph{Memory Construction and Reusable KV Composition.}
Long-term memory systems externalize prior interactions and decide what to
retain and retrieve. MemGPT~\cite{memgpt} manages memory tiers, while
Memory-R1~\cite{memoryr1} learns structured memory operations together with
answering. A complementary line reuses latent states: Prompt
Cache~\cite{promptcache} caches prompt modules; CacheBlend~\cite{cacheblend}
selectively recomputes tokens; Block-Attention~\cite{blockattention},
KVLink~\cite{kvlink}, and InferScale~\cite{inferscale} compose cached blocks
using blockwise attention, links, positional correction, or contextual
encoding. These systems establish efficient KV reuse, but their end-to-end
evaluations couple candidate selection, representation, and downstream use.
Recomputation and links can restore interactions, yet the final score does not
show whether an improvement came from a richer state or a better serving path,
nor where contextual fidelity begins to conflict with independent reuse.
EASE-KV fixes candidates and the backbone to determine whether a failure
originates in the encoded state or its later allocation.

\paragraph{Memory Influence Control and Outcome Supervision.}
PASTA, AutoPASTA, and Spotlight steer a frozen model toward selected prompt
regions~\cite{pasta,autopasta,spotlight}. KV Cache Steering and Memory
Inception intervene directly on latent states~\cite{kvsteering,memoryinception}.
These methods establish controllability, but do
not study continuous competition among retrieved memories separately from
representation fidelity. Outcome-based methods offer another route:
Memory-R1 optimizes memory decisions from task outcomes, while GRPO and its
refinements~\cite{grpo,dapo,drgrpo} learn from sequence-level rewards. A
single outcome, however, may assign the same reward to many memory-wise
actions. EASE-KV therefore separates whether a useful allocation exists from
whether the observed reward can identify a transferable policy. Unlike global
steering, its action is indexed by retrieved memory; unlike a new RL method,
its supervision analysis measures the distinctions exposed to a learner.
%========================================================

\section{Methodology}
\label{sec:method}

EASE-KV separates post-retrieval memory failures into three interfaces under
fixed candidates and a frozen decoder: \emph{encoding} determines which
relations survive in cached states, \emph{allocation} controls their relative
influence, and \emph{supervision} determines whether useful controls can be
learned. This decomposition distinguishes missing representation, underused
representation, and insufficiently informative supervision.

\subsection{Problem Formulation}
\label{sec:setting}

For query $q$, retrieval returns a candidate set
$M_q=\{m_1,\ldots,m_K\}$. Each $m_k$ is an atomic statement distilled from
a dialogue turn. Controlled experiments
ensure that candidates contain the required evidence, isolating post-retrieval
behavior. With candidates, prompt, decoding, and frozen model $F_\theta$
matched, the text channel serializes $M_q$, whereas the latent channel encodes
memories once and injects their layer-wise KV states:
\begin{equation}
M_q
\xrightarrow{E_\theta(\cdot;\mathcal C)}
\mathcal K_q
\xrightarrow{\mathbf b}
F_\theta(q,\mathcal K_q;\mathbf b)
\rightarrow y.
\label{eq:overall_pipeline}
\end{equation}
Here, $\mathcal C$ is the encoding context, $\mathcal K_q$ the resulting KV
cache, and $\mathbf b=(b_1,\ldots,b_K)$ the per-memory influence controls.
The encoder and answering model share frozen parameters; only an optional
allocation policy is trained. Text--latent comparisons test the delivery
format, encoding comparisons test preserved information, and allocation
comparisons test the use of fixed representations.

\begin{figure*}[t]
    \centering
    \includegraphics[width=0.90\textwidth]{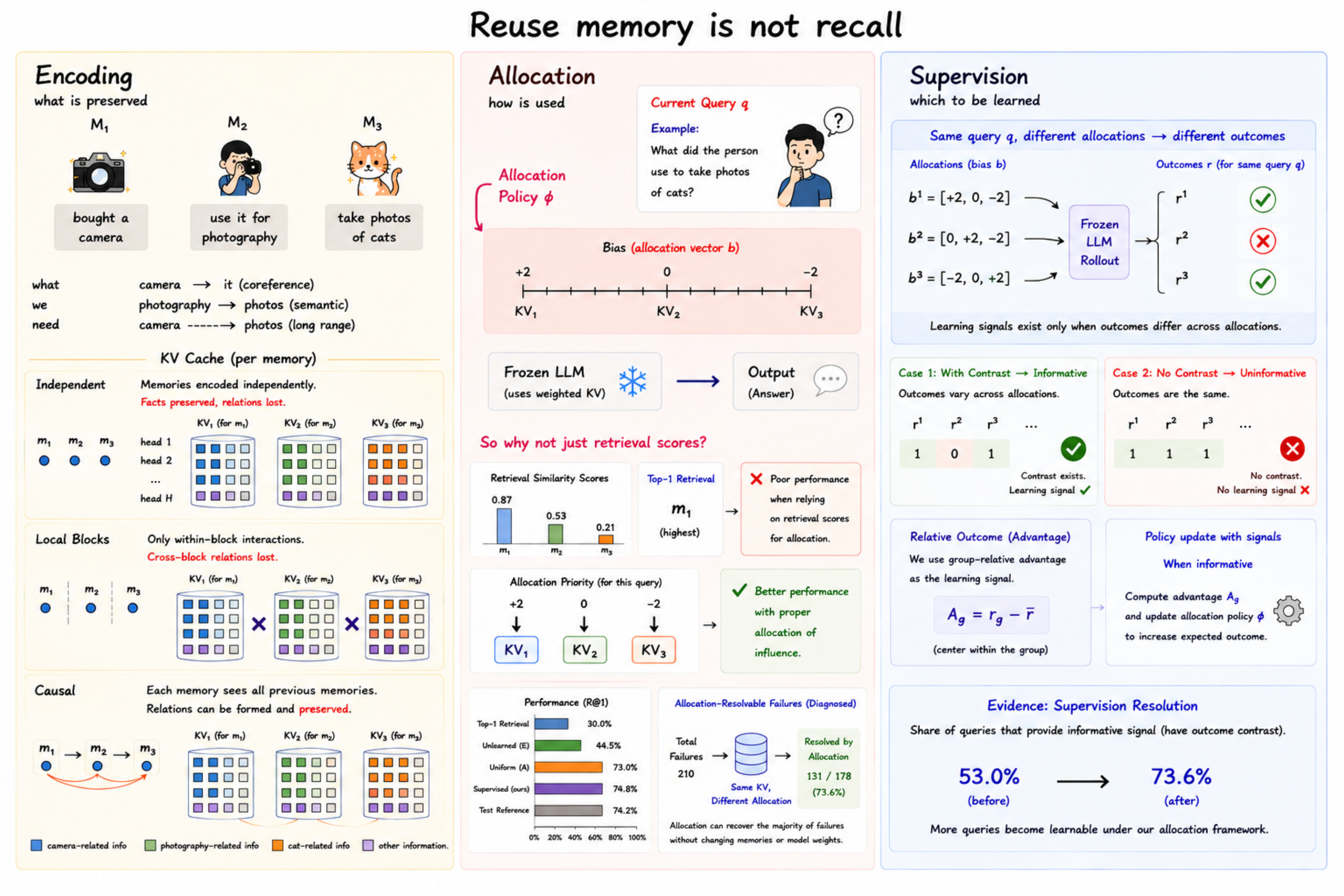}
    \caption{EASE-KV asks three sequential questions under matched candidates
    and a frozen model: what is encoded, what can be allocated, and what can
    be learned from outcome supervision.}
    \label{fig:easekv_overview}
\end{figure*}

\subsection{Encoding: What the Cache Preserves}
\label{sec:representation}

Encoding sequence $X$ under visibility context $\mathcal C$ produces
layer-wise keys and values:
\begin{equation}
E_\theta(X;\mathcal C)
=
\left\{(K_\ell,V_\ell)\right\}_{\ell=1}^{L}.
\label{eq:encoding}
\end{equation}
Identical tokens can yield different reusable states depending on which
memories were visible during encoding. We vary visibility while fixing memory
content and serving conditions.

\paragraph{Matched visibility.}
\emph{Independent} encoding processes each fact separately.
\emph{Block-diagonal} encoding places facts in one sequence but masks
cross-fact attention. \emph{Causal} encoding uses the same token layout and
positions as block-diagonal encoding while allowing later facts to attend
to earlier ones. The block-diagonal--causal comparison therefore isolates
cross-fact contextualization under matched positions, sequence length, and
fact boundaries. Independent encoding serves as an additional modular
reference but is not interchangeable with the matched block-diagonal
condition.

\paragraph{Reuse granularity.}
Fact-, turn-, and session-level caches package progressively more dialogue
context into each reusable unit. Query-level encoding jointly processes
$M_q$ and serves only as a contextualization reference because it is
query-dependent. These settings expose the trade-off between relational
fidelity and independent reuse: visibility tests interaction between facts,
whereas granularity tests the cost of storing their context together.

\subsection{Allocation: How the Cache Is Used}
\label{sec:injection}
\label{sec:allocation_interface}

We store keys before serving-time rotary embeddings, concatenate retrieved KV
banks with the query-side cache at each layer, and apply rotary positions at
the target splice locations. Subsequent prompt positions are shifted by the
inserted span. This correction aligns cached states with native prefill while
keeping the stored representations query-independent, preventing positional
mismatch from masquerading as semantic memory loss.

Let $g(j)$ map injected token $j$ to memory $m_k$. EASE-KV adds a scalar bias
$b_{g(j)}$ to its attention logit:
\begin{equation}
\widetilde z_{tj}^{(\ell)}
=
\frac{(q_t^{(\ell)})^\top k_j^{(\ell)}}{\sqrt{d_k}}
+
b_{g(j)}.
\label{eq:bias}
\end{equation}
Here, $q_t^{(\ell)}$ and $k_j^{(\ell)}$ are the query and key vectors at layer
$\ell$, and $d_k$ is the key dimension. The bias is shared by all injected
tokens from the same memory. Attention weights become
\begin{equation}
\widetilde\alpha_{tj}^{(\ell)}
=
\frac{\exp(\widetilde z_{tj}^{(\ell)})}
{\sum_u \exp(\widetilde z_{tu}^{(\ell)})}.
\label{eq:attention}
\end{equation}
Non-memory positions receive zero bias; positive and negative biases
respectively increase and decrease a memory's relative influence.

For two memory tokens $i$ and $j$ competing for the same query, the attention
odds satisfy
\begin{equation}
\frac{\widetilde\alpha_{ti}^{(\ell)}}
     {\widetilde\alpha_{tj}^{(\ell)}}
=
\exp(b_{g(i)}-b_{g(j)})
\frac{\alpha_{ti}^{(\ell)}}{\alpha_{tj}^{(\ell)}}.
\label{eq:relative_influence}
\end{equation}
Here, $\alpha$ and $\widetilde\alpha$ denote attention weights before and
after biasing. The intervention changes relative competition without
rewriting stored keys or values; the evaluated bias range is $[-4,4]$.

Candidates, KV tensors, positions, and decoder remain fixed in matched
allocation comparisons. Consequently, answer differences can be attributed
to the tested influence controls rather than changes to the stored
representations. This does not establish that the chosen control family
can recover all information lost during encoding.

We compare unsteered injection ($\mathbf b=\mathbf 0$), uniform bias, and the
oracle-evidence reference Uniform $\pm2$, which assigns $+2$ to evidence and
$-2$ to distractors. This oracle is a diagnostic reference, not a deployable
method. A policy $\pi_\phi(q,M_q)$ instead predicts one bias per candidate.
Uniform controls test broad amplification and suppression, while learned
policies test whether allocation decisions can generalize without observing
test outcomes. Outcome-aware per-query search, when used, is interpreted
only as a diagnostic control rather than a deployable allocation strategy.

\subsection{Supervision: What Allocation Can Be Learned}
\label{sec:policy}
\label{sec:learning}

For each query, the policy samples $G$ allocations:
\begin{equation}
\mathbf b^{(g)}
\sim
\pi_\phi(\cdot\mid q,M_q),
\qquad g=1,\ldots,G.
\label{eq:policy_sampling}
\end{equation}
The frozen model generates answer $y^{(g)}$ for each allocation. We write
the reward in its general form as
\begin{equation}
\begin{aligned}
r_g
&=R(y^{(g)},y^*,\mathbf b^{(g)}),\\
\bar r
&=\frac{1}{G}\sum_{g=1}^{G}r_g,
\end{aligned}
\label{eq:group_reward}
\end{equation}
where $y^*$ is the reference answer and $\bar r$ is the mean reward within
the query group. The reward may depend on both answer quality and
allocation-dependent shaping terms. In particular, binary correctness
used to diagnose reward contrast should not be conflated with the
complete reward used by a learned allocation policy.

The group-relative advantage is
\begin{equation}
A_g=r_g-\bar r.
\label{eq:group_advantage}
\end{equation}
At the supervision-interface level, a group-relative objective takes
the form
\begin{equation}
\mathcal L_{\mathrm{out}}
=
-\frac{1}{G}\sum_{g=1}^{G}
A_g\log
\pi_\phi(\mathbf b^{(g)}\mid q,M_q).
\label{eq:outcome_objective}
\end{equation}
This objective describes how above-average allocations can be favored
over below-average ones. It is used to analyze the supervision available
for memory-wise decisions rather than to introduce a new RL optimizer.
The complete allocation receives an answer-level evaluation, while
action-dependent reward terms may additionally shape the policy.

Evidence supervision labels memory relevance but does not observe the effect
of a bias. Outcome-based supervision evaluates the resulting allocation,
but its informativeness depends on whether different allocations sampled
for the same query receive distinguishable rewards.

We measure the fraction of queries with non-zero within-group reward
variance:
\begin{equation}
\rho_{\mathrm{inform}}
=
\frac{1}{N}\sum_{q=1}^{N}
\mathbf 1
\left[
\operatorname{Var}_{g}(r_{q,g})>0
\right].
\label{eq:informative_fraction_method}
\end{equation}
Here, $N$ is the number of queries, $r_{q,g}$ is the reward of allocation
$g$ for query $q$, and $\mathbf 1[\cdot]$ indicates non-zero variance.

Under binary correctness, groups in which all sampled allocations succeed
or all fail have zero reward variance and provide no within-group
preference. Shaped rewards can distinguish allocations that receive the
same binary outcome, although reward variation alone does not establish
that the resulting preference is useful or transferable.

We therefore distinguish reward-resolution diagnostics from policy-learning
results. Non-zero reward variance is necessary for group-relative learning
under the corresponding reward, but it is insufficient to establish
generalization: a revealed preference may not be predictable from
$(q,M_q)$. Held-out policy evaluation tests whether the learned allocation
rule transfers to unseen queries.

\subsection{Matched Evaluation Protocol}
\label{sec:calibration}

Dialogue turns are distilled into atomic statements while preserving names,
dates, quantities, and task-relevant attributes. To distinguish content
loss during distillation from failures of latent KV reuse, text-rendered
memories are evaluated separately from their injected KV representations.
A text-based diagnostic does not by itself establish the fidelity of the
stored KV states.

The block-diagonal--causal comparison fixes contents, layout, injection, and
allocation; allocation comparisons reuse identical KV states; supervision
comparisons additionally control policy architecture, decoding, and
evaluation settings. Before semantic evaluation, native prefill and
injection are calibrated to a bf16 numerical noise floor using layer-wise
states, query representations, logits, and greedy generations after position
correction. This checks implementation equivalence, not predictive
accuracy. Decoding is otherwise greedy, with a shared answer judge.

Evaluation subsets constructed using reference-answer correctness are
treated as diagnostic subsets rather than independently held-out test
sets. Such selection must be distinguished from text-fidelity validation
and from evaluation on an unfiltered benchmark. Comparisons are interpreted
within their respective fixed evaluation conditions rather than pooled
across differently selected subsets.

Outcome-based search used to construct policy-training targets is
distinguished from per-query diagnostic search. The latter may inspect
outcomes for the evaluated query and therefore cannot establish deployable
test-time policy performance. Learned allocation policies are assessed
separately on held-out queries without per-query outcome search.

\section{Experiments}
\label{sec:experiments}

We evaluate EASE-KV in causal order: encoding experiments test whether the
required relations survive cache construction; allocation experiments fix
those states and measure the effect of controlling memory influence;
supervision experiments test whether useful allocation decisions transfer
to held-out queries.

\begin{figure*}[t]
    \centering
    \includegraphics[width=\textwidth]{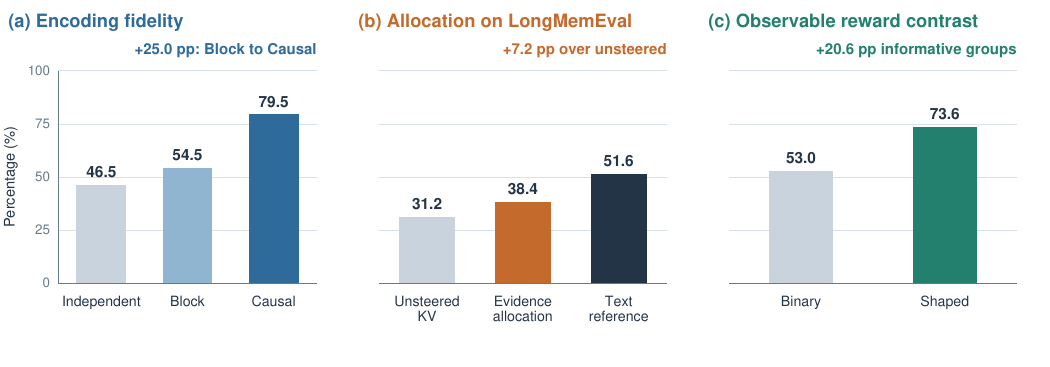}
    \caption{Headline evidence for encoding, allocation, and supervision.
    Panels report separate controlled experiments and are compared only
    within panel.}
    \label{fig:easekv_results}
\end{figure*}

\subsection{Experimental Setup}
\label{sec:experimental_setup}

\paragraph{Setup.}
Our controlled compositional setting fixes candidates containing the required
evidence and separates single- from multi-hop questions.
LongMemEval~\cite{longmemeval} provides retrieved memories from long
conversations; a full-haystack variant adds distractors, and a temporal-conflict
setting retains old and new facts while querying the current state.
Main experiments use frozen Qwen2.5-7B-Instruct~\cite{qwen25}; encoding is
repeated with Llama 8B and Gemma 12B. Candidates, greedy decoding, and
the semantic-equivalence judge are fixed within each comparison.

These settings play complementary roles. The controlled task supports
causal attribution because required evidence is guaranteed to be available.
LongMemEval tests the interfaces under a long-conversation retrieval
distribution, while full-haystack and temporal conflict stress irrelevant
and contradictory memories. We do not aggregate them because each isolates
a different failure.

\paragraph{References and metrics.}
Text is a delivery reference rather than an efficiency-matched baseline.
Unsteered KV uses $\mathbf b=\mathbf 0$; Uniform $\pm2$ and hand-specified
recency are diagnostic controls because they use annotations or information
not necessarily available to a deployed policy. Outcome-aware per-query
search, when considered, is also diagnostic because it observes evaluated
outcomes. We report accuracy and percentage-point differences, plus
evidence-selection precision, recall, and PR-AUC where applicable.
Reward-resolution experiments additionally report $\rho_{\mathrm{inform}}$,
the fraction of query groups with non-zero reward variance.

\paragraph{Evidence hierarchy.}
Matched visibility comparisons identify the effect of encoding-time
contextualization. Oracle-guided allocation measures the influence
available through fixed KV states, whereas held-out learned policies
evaluate whether allocation decisions generalize. Improvements under
oracle control should not be interpreted as deployable policy performance.

\subsection{Encoding Fidelity}
\label{sec:encoding}

\begin{table}[t]
    \centering
    \scriptsize
    \setlength{\tabcolsep}{3pt}
    \renewcommand{\arraystretch}{1.05}
    \begin{tabular}{@{}lrrr@{}}
        \toprule
        \rowcolor{black!4}
        \textbf{Backbone} & \textbf{Block-diag.}
        & \textbf{Causal} & \textbf{Gain} \\
        \midrule
        Qwen 7B & 54.5 & \best{79.5} & +25.0 \\
        Llama 8B & 64.0 & \best{74.0} & +10.0 \\
        Gemma 12B & 69.0 & \best{79.0} & +10.0 \\
        \bottomrule
    \end{tabular}
    \caption{Cross-backbone encoding accuracy (\%). Block-diagonal and
    causal encoding share the token layout and positions; their difference
    measures the effect of cross-fact visibility.}
    \label{tab:cross_model}
\end{table}

\paragraph{Cross-memory visibility.}
Independent, block-diagonal, and causal encoding share memory content,
injection, and allocation. Block-diagonal and causal encoding additionally
share token layout, positions, and fact boundaries; only cross-fact
visibility differs between this matched pair.
Figure~\ref{fig:easekv_results}(a) reports $46.5\%$, $54.5\%$, and
$79.5\%$ accuracy for these three conditions, respectively.
The $25.0$-point causal gain over the matched block-diagonal condition
therefore isolates the contribution of encoding-time cross-fact visibility
under the evaluated setup. Retrieving all facts is insufficient when
independently cached states fail to preserve the relations needed to
compose them.

Independent encoding achieves $46.5\%$, below the $54.5\%$ block-diagonal
condition. However, independent encoding also changes the encoding
procedure and layout, so this difference is not interpreted as a pure
visibility effect. The block-diagonal--causal comparison provides the
matched evidence for contextualization.

\paragraph{Cross-backbone consistency.}
Table~\ref{tab:cross_model} reports the matched block-diagonal--causal
comparison across three model families. Causal encoding improves accuracy
by $25.0$ percentage points for Qwen and by $10.0$ points for both
Llama and Gemma. Although the magnitude varies, all three tested
backbones show a positive effect of cross-fact visibility under their
respective matched conditions.

\subsection{Allocation of Preserved Memory}
\label{sec:allocation}

\begin{table}[t]
    \centering
    \scriptsize
    \setlength{\tabcolsep}{3pt}
    \renewcommand{\arraystretch}{1.05}
    \begin{tabular}{@{}lrrrr@{}}
        \toprule
        \rowcolor{black!4}
        \textbf{Setting} & \textbf{Text} & \textbf{Latent}
        & \textbf{Diag.} & \textbf{Learned} \\
        \midrule
        LongMemEval & \best{51.6} & 31.2 & 38.4 & -- \\
        Full-haystack & \best{55.5} & 38.6 & 41.5 & -- \\
        Temporal & \best{77.8} & 56.9 & 76.4 & 70.8 \\
        \bottomrule
    \end{tabular}
    \caption{End-to-end accuracy (\%). Diagnostic allocation uses
    oracle evidence or temporal information and is not a deployable
    policy result.}
    \label{tab:end_to_end}
\end{table}

\paragraph{Memory competition.}
On LongMemEval, diagnostic evidence-guided allocation raises latent
accuracy from $31.2\%$ to $38.4\%$ with unchanged candidates, KV states,
and backbone. The $7.2$-point gain demonstrates that changing memory
influence can improve accuracy without re-encoding, while leaving a
$13.2$-point gap to the text reference. Under additional distractors,
the corresponding gain is $2.9$ points ($38.6\%$ to $41.5\%$).

In temporal conflict, hand-specified recency reaches $76.4\%$, compared
with $77.8\%$ for timestamped text and $56.9\%$ for the latent reference;
the learned allocation policy reaches $70.8\%$. These results indicate
that the evaluated allocation interface can express a preference for
the updated memory without deleting the earlier one. However,
hand-specified recency relies on temporal information and is therefore
a diagnostic control rather than evidence of learned temporal selection.

Across these settings, allocation changes the answer while stored KV
states remain fixed. The observed improvements measure control over
existing representations rather than gains from new encoding.
Their magnitude differs because the candidate sets and stored
representations differ across tasks.

\paragraph{Recovery profiles.}
For each setting, we divide the diagnostic allocation gain by the
text--latent gap. Evidence-guided allocation recovers
$7.2/20.4=35.3\%$ of the gap on LongMemEval and
$2.9/16.9=17.2\%$ under full-haystack distractors, leaving residual
gaps of $13.2$ and $14.0$ points, respectively.

Hand-specified recency instead recovers
$19.5/20.9=93.3\%$ of the temporal-conflict gap, leaving a
$1.4$-point residual. This result is consistent with the competing
memory versions remaining accessible through the fixed cache under
an appropriate temporal preference. By contrast, the larger
residuals in LongMemEval and full-haystack indicate that the tested
allocation controls do not close the full delivery gap.

These recovery profiles distinguish improvements obtained by
redistributing attention from remaining limitations of the
evaluated latent-memory interface. They do not establish that
the tested biases are optimal or that every residual error
originates from missing encoded information.

\subsection{Learning Allocation From Outcomes}
\label{sec:supervision}

\paragraph{Reward resolution.}
Figure~\ref{fig:easekv_results}(c) reports that binary correctness
provides non-zero within-query reward contrast for $53.0\%$ of
sampled query groups, compared with $73.6\%$ under the evaluated
shaped-reward setting. This comparison measures the distinguishability
of sampled allocations under different reward definitions, rather
than answer accuracy or policy generalization.

Under greedy decoding, distinct attention patterns and token
probabilities can produce the same final answer. Consequently,
binary correctness may assign identical rewards to allocations
that differ in their influence on the model. Reward shaping can
increase within-group variation, but a non-zero reward difference
does not by itself establish that the induced preference is useful
for learning a transferable policy.

\begin{table*}[t]
    \centering
    \small
    \setlength{\tabcolsep}{8pt}
    \renewcommand{\arraystretch}{1.08}
    \begin{tabular}{@{}llrr@{}}
        \toprule
        \rowcolor{black!4}
        \textbf{Regime} & \textbf{Method}
        & \textbf{Test (\%)} & \textbf{Gain (pp)} \\
        \midrule
        \multirow{2}{*}{Weak base}
          & Uniform reference & 48.0 & 0.0 \\
          & Outcome RL & \best{52.5} & +4.5 \\
        \addlinespace[2pt]
        \multirow{3}{*}{Strong base}
          & Uniform $\pm2$ & 75.0 & 0.0 \\
          & Warm-neutral RL & \best{75.5} & +0.5 \\
          & From-scratch RL & 73.5 & -1.5 \\
        \addlinespace[2pt]
        \multirow{3}{*}{Search--distill}
          & Uniform $\pm2$ & \best{75.0} & 0.0 \\
          & Failure-only targets & 66.5 & -8.5 \\
          & Mixed targets & 74.0 & -1.0 \\
        \bottomrule
    \end{tabular}
    \caption{Descriptive accuracy of learned and distilled allocations
    against their matched references. Gains are percentage-point
    differences; no statistical significance is inferred from these
    comparisons.}
    \label{tab:learning_main}
\end{table*}

\paragraph{Reward contrast and policy transfer.}
Outcome-guided RL improves the weak reference from $48.0\%$ to
$52.5\%$, a descriptive gain of $4.5$ points. Against the stronger
Uniform $\pm2$ reference at $75.0\%$, warm-neutral RL reaches
$75.5\%$ and from-scratch RL reaches $73.5\%$. Thus, the evaluated
learned policies do not exhibit a consistent descriptive improvement
over the strong evidence-guided reference.

Search--distillation provides a complementary learning setting:
allocation targets are constructed from training-side searches and
then used to train a policy evaluated on held-out queries.
Failure-only targets yield $66.5\%$ accuracy, while mixed targets
reach $74.0\%$, compared with $75.0\%$ for Uniform $\pm2$.
These comparisons suggest that access to outcome-derived targets
does not automatically produce a better transferable allocation
rule under the evaluated training conditions.

Together, the reward-resolution and policy-learning results
distinguish observable feedback from generalizable control.
More informative rewards can expose differences between sampled
allocations, but the learned policy must still predict useful
memory-wise decisions from the query and candidate set.

\section{Ablation Studies}
\label{sec:ablations}

We vary choices within each interface while retaining matched candidates
and a frozen backbone, isolating their contributions to memory use.

\subsection{Encoding Context and Reuse Granularity}
\label{sec:encoding_ablations}

\begin{table}[t]
    \centering
    \scriptsize
    \setlength{\tabcolsep}{3pt}
    \renewcommand{\arraystretch}{1.0}
    \begin{tabular}{@{}lrr@{}}
        \toprule
        \rowcolor{black!4}
        \textbf{Condition} & \textbf{Acc.} & \textbf{$\Delta$ (pp)} \\
        \midrule
        \rowcolor{black!2}
        \multicolumn{3}{@{}l}{\textit{Visibility}} \\
        Independent & 46.5 & -33.0 \\
        Block-diagonal & 54.5 & -25.0 \\
        Causal & \best{79.5} & 0.0 \\
        \addlinespace[1pt]
        \rowcolor{black!2}
        \multicolumn{3}{@{}l}{\textit{Inference bridge}} \\
        Text reference & 31.2 & -0.8 \\
        Turn-level KV & 11.2 & -20.8 \\
        Session-level KV & \best{32.0} & 0.0 \\
        Separator & 10.4 & -21.6 \\
        Indexed anchor & 9.6 & -22.4 \\
        Semantic anchor & 11.2 & -20.8 \\
        \bottomrule
    \end{tabular}
    \caption{Encoding ablations. Visibility differences use causal
    encoding as reference; bridge differences use session-level KV.}
    \label{tab:encoding_ablations}
\end{table}

\paragraph{The gain comes from encoding-time visibility.}
Table~\ref{tab:encoding_ablations} compares encoding conditions and
inference-time bridges. Block-diagonal and causal encoding share token
layout and positions but differ in cross-fact attention; the latter gains
$25.0$ points. Independent encoding is a separate modular reference.
Conversely, separators, indices, and semantic anchors added after retrieval
do not improve the $11.2\%$ turn-level reference and remain $21$--$22$
points below session-level encoding. These tested serving-time cues therefore
fail to recover the contextual information lost during independent caching.

\paragraph{Context and reuse.}
Session-level KV reaches $32.0\%$, comparable to the $31.2\%$ text
reference in the bridge study, while turn-level KV remains at $11.2\%$.
Broader encoding context can preserve cross-turn relations, but the session
then becomes one retrieval unit. Fact- and turn-level caches allow finer
reuse, whereas query-level encoding offers greater contextualization at
the cost of query dependence. The results expose a context--reuse trade-off
rather than a universal benefit from larger units.

\subsection{Allocation Mechanisms}
\label{sec:allocation_ablations}

\begin{table}[t]
    \centering
    \scriptsize
    \setlength{\tabcolsep}{3pt}
    \renewcommand{\arraystretch}{1.05}
    \begin{tabular}{@{}lrrr@{}}
        \toprule
        \rowcolor{black!4}
        \textbf{Allocation} & \textbf{Overall} & \textbf{1-hop}
        & \textbf{Multi-hop} \\
        \midrule
        Text ref. & 74.5 & 77.9 & \best{64.7} \\
        Unsteered & 44.5 & 51.0 & 25.5 \\
        Sim. top-1 & 30.0 & 33.6 & 19.6 \\
        Uniform $\pm2$ & \best{75.0} & \best{83.9} & 49.0 \\
        \bottomrule
    \end{tabular}
    \caption{Allocation accuracy (\%). Uniform $\pm2$ uses oracle
    evidence labels and serves as a diagnostic reference.}
    \label{tab:allocation_main}
\end{table}

\paragraph{Retrieval relevance is not post-retrieval allocation.}
Amplifying the top similarity match lowers overall accuracy from $44.5\%$
to $30.0\%$ and multi-hop accuracy from $25.5\%$ to $19.6\%$:
the highest direct match need not be the bridge fact. Oracle-guided
Uniform $\pm2$ instead reaches $75.0\%$ overall and $83.9\%$ on
single-hop questions, but only $49.0\%$ on multi-hop, below the text
reference at $64.7\%$. These results show that evidence membership
can guide memory competition, but retrieval similarity alone is not
a reliable allocation rule.

\paragraph{Selection and allocation.}
The evidence selector achieves conversation-disjoint precision of
$64.50\%$, recall of $46.02\%$, and PR-AUC of $60.62\%$, with
robustness variants spanning $60.25$--$69.4\%$ PR-AUC.
These measurements characterize evidence identification rather than
the downstream effect of signed attention biases. Predicting
relevance and determining effective memory influence are distinct
problems.

\FloatBarrier
\subsection{Supervision Resolution and Policy Priors}
\label{sec:supervision_ablations}

\begin{table}[t]
    \centering
    \small
    \setlength{\tabcolsep}{4pt}
    \renewcommand{\arraystretch}{1.08}
    \begin{tabularx}{\columnwidth}{@{}lrr>{\raggedright\arraybackslash}X@{}}
        \toprule
        \rowcolor{black!4}
        \textbf{Initialization} & \textbf{Dev (\%)} & \textbf{Test (\%)}
        & \textbf{Dominant action} \\
        \midrule
        Warm-neutral & 75.0 & \best{75.5}
        & $85\%$ at $0$; $14\%$ at $+2$ \\
        From scratch & 73.2 & 73.5
        & $81\%$ at $-4$ \\
        Warm-legacy & 64.3 & --
        & $46\%$ at $-4$; $32\%$ at $-2$ \\
        Uniform $\pm2$ & -- & 75.0
        & Fixed reference \\
        \bottomrule
    \end{tabularx}
    \caption{Initialization under sparse within-query reward contrast.}
    \label{tab:initialization}
\end{table}

\paragraph{Reward contrast and policy priors.}
The reported reward-resolution comparison shows that shaping increases
informative groups from $53.0\%$ to $73.6\%$; a separate audit associates
$27.9\%$ of shaping pseudo-gradient updates with informative contrast.
Initialization also affects the learned action distribution: warm-neutral
places $85\%$ of actions at zero and reaches $75.5\%$, whereas training
from scratch places $81\%$ at $-4$ and reaches $73.5\%$.
Similar accuracy can therefore mask different allocation policies.

\paragraph{The learning bottleneck is not only exploration.}
RL adds $4.5$ points over a weak reference but only $0.5$ over the
strong Uniform $\pm2$ reference. Search--distill removes on-policy
exploration, yet failure-only and mixed targets remain $8.5$ and
$1.0$ points below Uniform $\pm2$, respectively. These descriptive
results suggest that outcome-derived targets do not automatically
yield transferable allocation rules. Sparse reward contrast and
initialization-dependent behavior further complicate policy learning.

\section{Discussion}
\label{sec:discussion}

EASE-KV reframes reusable memory as a sequence of three conditions rather than
one end-to-end score. Encoding determines which relations survive cache
construction; allocation determines how preserved memories compete; and
supervision determines whether useful controls can be identified. The matched
results---a $25.0$-point gain from causal visibility, a $7.2$-point gain from
diagnostic evidence-guided allocation on LongMemEval, and informative groups
increasing from $53.0\%$ to $73.6\%$ under shaped reward---illustrate
distinct constraints at these interfaces. The resulting design rule is to
preserve relations before optimizing influence and verify reward resolution
before blaming the allocator. Otherwise, aggregate accuracy conflates
representation fidelity, controllability, and policy learnability.

KV-memory systems can be evaluated in three steps: first test contextual
visibility, then measure allocation headroom with the state fixed, and finally
compare that headroom with within-query reward contrast before judging a
learned policy. This order separates gains from richer representations from
improved access to existing information.

The same decomposition gives a modular design principle. Cache construction
should preserve required relations within reusable context; retrieval
determines candidate availability; allocation resolves query-dependent
competition; and supervision distinguishes alternatives. Retrieval rank
therefore need not determine attention priority, while increased policy
capacity alone cannot resolve indistinguishable rewards.

\section{Conclusion}
\label{sec:conclusion}

Reuse is not recall. Effective cached memory must preserve required relations,
allocate appropriate influence, and provide supervision that distinguishes
better memory use. EASE-KV evaluates these conditions through matched encoding
comparisons, fixed-state allocation controls, and reward-resolution analysis.
This decomposition helps identify whether improvements are needed in encoding,
allocation, or supervision rather than treating every failure as a retrieval
problem.

\section{Limitations}
\label{sec:limitations}

We fix retrieved candidates and the language model, isolating post-retrieval
memory use rather than joint retriever--allocator optimization. Oracle-evidence
allocation and hand-specified recency are diagnostic controls, not deployable
policies. The evaluated results may vary across backbones, memory distributions,
and allocation settings. Reward contrast measures distinguishability but does
not guarantee policy generalization. We establish a diagnostic decomposition,
not the optimality of the tested encoding strategies, allocation biases, or
learning objectives.

% References
% acl.sty selects acl_natbib automatically.

% References
\bibliography{thereference}

% Appendix
\clearpage
\appendix
\section{Datasets and Task Construction}
\label{app:data}

The experiments comprise distinct controlled comparisons rather than a
single pooled benchmark. Compositional questions isolate memory encoding
and allocation with known evidence; LongMemEval examines delivery under
long conversational histories; full-haystack and temporal-conflict variants
stress distractors and updates. A separate LoCoMo-based analysis evaluates
evidence selection. Table~\ref{tab:app_datasets} distinguishes their purposes.
We report sample counts only for evaluation files and splits for which the
underlying records have been identified; counts from one experiment are not
assumed to apply to the others.

\begin{table*}[t]
    \centering
    \small
    \setlength{\tabcolsep}{4.5pt}
    \renewcommand{\arraystretch}{1.08}
    \begin{tabularx}{\textwidth}{@{}>{\raggedright\arraybackslash}p{2.6cm}>{\raggedright\arraybackslash}p{4.0cm}>{\raggedright\arraybackslash}X@{}}
        \toprule
        \rowcolor{black!4}
        \textbf{Setting} & \textbf{Candidate condition} & \textbf{Primary role} \\
        \midrule
        Controlled composition
        & Required facts are present; single- and multi-hop questions are separated
        & Isolate cross-memory encoding visibility and post-retrieval allocation. \\
        LongMemEval
        & Retrieved memories are fixed within each text--latent comparison
        & Measure representation loss and recoverable allocation headroom under long conversational histories. \\
        Full-haystack
        & The candidate set contains additional irrelevant memories
        & Stress competition between evidence and distractors. \\
        Temporal conflict
        & Earlier and updated facts remain simultaneously available
        & Test whether allocation can express a recency preference without deleting history. \\
        LoCoMo selector audit
        & Conversation-disjoint evidence-selection evaluation
        & Test whether memory relevance can be predicted without conflating selection with answer accuracy. \\
        \bottomrule
    \end{tabularx}

    \caption{Evaluation settings and the interface each one is designed to
    expose. Candidate control describes the comparison used in this paper,
    not a property of the original benchmark.}
    \label{tab:app_datasets}
\end{table*}

\subsection{Controlled Compositional Setting}

The controlled setting is constructed around atomic memories whose required
evidence is included in the candidate set. A single-hop question can be
answered from one memory. A multi-hop question requires a chain of at least
two memories, so no individual candidate is sufficient. This separation is used in
Table~\ref{tab:allocation_main} to test whether similarity-based allocation
can overlook a bridge fact.

The setting is not intended to estimate retrieval recall. Its purpose is to
hold retrieval success constant so that a wrong answer can be attributed to
the representation delivered by the cache or to the influence assigned to
the delivered memories. Evidence and distractor turns undergo the same
distillation and encoding procedure. The verified evaluation records
underlying the four retained rows of Table~\ref{tab:allocation_main} contain
200 questions, comprising 149 single-hop and 51 multi-hop questions. All four
methods use the same question identifiers. Table~\ref{tab:app_allocation_counts}
reports their integer correctness counts, so the displayed percentages can
be independently checked.

\begin{table}[t]
    \centering
    \scriptsize
    \setlength{\tabcolsep}{3.5pt}
    \renewcommand{\arraystretch}{1.05}
    \begin{tabular}{@{}lrrr@{}}
        \toprule
        \rowcolor{black!4}
        \textbf{Allocation} & \textbf{All} & \textbf{1-hop}
        & \textbf{Multi-hop} \\
        \midrule
        Text reference & 149/200 & 116/149 & 33/51 \\
        Unsteered KV & 89/200 & 76/149 & 13/51 \\
        Similarity top-1 & 60/200 & 50/149 & 10/51 \\
        Uniform $\pm2$ & 150/200 & 125/149 & 25/51 \\
        \bottomrule
    \end{tabular}
    \caption{Verified correct/total counts for the four retained
    allocation-ablation rows in Table~\ref{tab:allocation_main}. These are
    correctness counts, not counts of evidence memories.}
    \label{tab:app_allocation_counts}
\end{table}

\subsection{LongMemEval}

The LongMemEval benchmark contains 500 curated questions embedded in
scalable, timestamped user--assistant histories and covers information
extraction, multi-session reasoning, temporal reasoning, knowledge updates,
and abstention~\cite{longmemeval}. This is the benchmark's nominal size;
it must not be treated as the verified denominator of every LongMemEval
variant without a run-specific evaluation manifest. We use the benchmark
to compare text delivery, unsteered latent KV, and evidence-guided allocation.
The retrieved candidates, prompt, decoder, and judge are fixed
within each comparison. Thus the text--latent difference measures a delivery
gap under the selected memories, while the difference between unsteered and
allocated KV measures how much of that gap can be changed without re-encoding.

We do not aggregate LongMemEval with the controlled task. LongMemEval mixes
memory abilities and a realistic retrieval distribution, whereas the
controlled task identifies a particular causal interface. Their scores
therefore answer different questions.

\subsection{Full-Haystack and Temporal-Conflict Variants}

The full-haystack variant retains the same delivery comparison but increases
the number of irrelevant memories competing with the evidence. It tests an
important boundary of allocation: biases can redistribute influence among
available candidates, but cannot replace missing evidence or repair an
unfavorable candidate set.

The temporal-conflict setting contains an earlier memory
$m_{\mathrm{old}}$ and a later update $m_{\mathrm{new}}$. Queries ask for the
current state while both memories remain in the cache. Timestamped text is
the delivery reference; hand-specified recency is a diagnostic latent
control; and outcome RL tests whether the preference can be learned. The
setting therefore separates control expressivity from policy recovery.

\subsection{LoCoMo Evidence-Selector Audit}

LoCoMo consists of long conversations spanning multiple sessions and
supports evaluation of long-term conversational memory~\cite{locomo}.
We use conversation-disjoint evaluation for the evidence-selector audit.
The seven-fold protocol holds out one conversation per fold, evaluating
relevance prediction on conversations excluded from the corresponding
selector fit. Evidence-selection precision, recall, and PR-AUC are distinct
from downstream answer accuracy and do not measure the efficacy of the
selected memories' signed attention biases.

\subsection{Answer-Conditioned Diagnostic Subsets}

A separately audited ceiling gate retained questions for which the
text-oracle answer was judged correct on the evaluated test set. It reduced
200 evaluated questions to 149 and excluded 51. Because inclusion depends
on test-reference correctness, this is an \emph{answer-conditioned diagnostic
subset}, not an independently selected held-out benchmark. The subset's
149 retained questions must not be confused with the 149 single-hop
questions in the allocation-ablation evaluation; the matching count does
not establish matching identifiers or criteria.

This separate diagnostic audit records a pool of 750 training-side
questions, partitioned into 617 training and 133 development examples, and
a separate 200-question test evaluation. Its conversation-disjoint split
uses one development conversation and three test conversations. These counts
are specific to that audit and cannot be generalized to the encoding,
LongMemEval, or policy-training experiments. This answer-conditioned gate is not established to be the same as any
text-fidelity validation procedure. No verified implementation of a 95\%
text-fidelity inclusion threshold is available. We therefore do
not use the threshold as a methodological inclusion rule or report accuracy
on the answer-conditioned subset as full-test performance.

\FloatBarrier
\section{Baselines and Diagnostic Controls}
\label{app:baselines}

EASE-KV is a causal decomposition rather than an efficiency leaderboard.
Accordingly, its empirical baselines are matched delivery, encoding,
allocation, and supervision conditions. Prior KV-serving systems motivate the interfaces but are not reimplemented as unmatched end-to-end baselines. Tables~\ref{tab:app_encoding_baselines}
and~\ref{tab:app_allocation_baselines} state exactly what each reported
condition changes and what conclusion it supports.

\paragraph{Weak and strong references.}
The weak-reference comparison starts from $48.0\%$ accuracy, while the
strong-reference comparison uses oracle-guided Uniform $\pm2$ at $75.0\%$.
These values refer to different comparisons and should not be pooled.
Descriptive improvement over a weak reference does not establish statistical
significance, and the strong reference uses evidence labels unavailable to
an ordinary deployed selector.

\paragraph{Delivery and encoding controls.}
The text reference establishes how much of the task is solvable when memory
content is delivered through the backbone's native token interface. It is a
semantic reference rather than a latency-matched baseline, because the text
must be processed again for every query. Independent, block-diagonal, and causal encoding hold memory content fixed
but need not match every implementation detail. Independent encoding processes
facts separately; its bare and scaffold variants are distinct configurations
and should not be merged into a single matched baseline. Block-diagonal and
causal encoding use a shared packed layout and positions, differing in
cross-fact visibility. Therefore only the latter comparison directly isolates
visibility under a matched layout; independent encoding remains a modular
reference. Causal encoding permits later facts to contextualize earlier
facts at the cost of encoding the chosen facts together.
Turn-, session-, and query-level caches extend this comparison along the same
axis. Inference bridges are applied only after retrieval, so any recovery from
separators, indices, or semantic anchors is attributable to serving-time
structure rather than re-encoding the memory.

\paragraph{Allocation controls.}
Allocation baselines operate on a fixed candidate cache. Unsteered KV assigns
no memory-specific bias and reveals the behavior induced by the stored states
alone. Similarity top-1 tests the common shortcut of converting retrieval rank
directly into attention priority; it can select a relevant item but cannot
express graded competition among several candidates. Uniform $\pm2$ uses
oracle evidence membership to test whether the injection interface can realize
a useful allocation when the relevant memories are known. It is consequently
a diagnostic reference, not a deployable selector. The recency condition
plays the same diagnostic role for temporal conflicts: it asks whether a
known ordering preference can be expressed through the allocation interface.

\paragraph{Learned and outcome-aware controls.}
An evidence-supervised interaction selector predicts relevance using the
candidate set; its relevance metrics are reported independently from
end-to-end allocation accuracy. Outcome-based policy learning uses answer
feedback and may also include action-dependent shaping or regularization.
Search--distill uses training-side searched targets followed by held-out
policy evaluation. In contrast, outcome-aware search on the query being
answered may inspect its judged result and is only an exploratory diagnostic;
no unverified search-rescue rate is used to support the paper's conclusions.

\begin{table*}[t]
    \centering
    \small
    \setlength{\tabcolsep}{4.5pt}
    \renewcommand{\arraystretch}{1.08}
    \begin{tabularx}{\textwidth}{@{}>{\raggedright\arraybackslash}p{2.8cm}>{\raggedright\arraybackslash}p{4.1cm}>{\raggedright\arraybackslash}X@{}}
        \toprule
        \rowcolor{black!4}
        \textbf{Condition} & \textbf{What changes} & \textbf{Interpretation} \\
        \midrule
        Text reference
        & Memories are serialized into the normal prompt
        & Delivery reference; not compute-matched to reusable KV. \\
        Independent KV
        & Each fact is encoded in a separate forward pass
        & Maximum modularity with no cross-fact encoding context. \\
        Block-diagonal KV
        & Facts share positions and sequence length, but cross-fact attention is masked
        & Matched layout control for causal visibility. \\
        Causal KV
        & Later facts can attend to earlier facts during cache construction
        & Measures the effect of cross-fact contextualization. \\
        Turn/session KV
        & Progressively more source context is bound into one stored unit
        & Measures the fidelity--reuse trade-off. \\
        Query-level KV
        & Current candidates are encoded jointly for the evaluated query
        & Contextualization reference; not a query-independent cache. \\
        Inference bridges
        & Separators, indices, or semantic anchors are added after retrieval
        & Test whether serving cues recover context absent from stored states. \\
        \bottomrule
    \end{tabularx}

    \caption{Delivery and encoding baselines. Conditions in one comparison
    share memory contents, prompt, frozen backbone, decoding, and evaluator.}
    \label{tab:app_encoding_baselines}
\end{table*}

\begin{table*}[t]
    \centering
    \small
    \setlength{\tabcolsep}{4.5pt}
    \renewcommand{\arraystretch}{1.08}
    \begin{tabularx}{\textwidth}{@{}>{\raggedright\arraybackslash}p{2.8cm}>{\raggedright\arraybackslash}p{4.1cm}>{\raggedright\arraybackslash}X@{}}
        \toprule
        \rowcolor{black!4}
        \textbf{Condition} & \textbf{Information used} & \textbf{Role} \\
        \midrule
        Unsteered KV
        & $\mathbf b=\mathbf0$
        & Fixed-representation reference. \\
        Similarity top-1
        & Highest retrieval-similarity candidate
        & Tests whether a retrieval rank is also a useful allocation rule. \\
        Uniform $\pm2$
        & Oracle evidence labels
        & Strong diagnostic reference: evidence gets $+2$, distractors $-2$. \\
        Evidence-supervised selector
        & Training evidence labels and the candidate set
        & Measures relevance prediction separately from answer accuracy. \\
        Outcome RL
        & Training-query outcomes and possible action-dependent shaping
        & Tests whether a learned allocation rule transfers to held-out queries. \\
        Search--distill
        & Searched targets from training or development queries
        & Removes on-policy exploration and tests target transfer. \\
        Hand-specified recency
        & Known temporal order
        & Diagnostic temporal expressivity control. \\
        \bottomrule
    \end{tabularx}

    \caption{Allocation and supervision baselines. Oracle annotations or
    evaluated outcomes make a condition diagnostic rather than deployable.}
    \label{tab:app_allocation_baselines}
\end{table*}

\FloatBarrier
\section{Memory Construction and Encoding Details}
\label{app:method_details}

\subsection{Atomic Memory Construction and Text-Channel Checks}

Raw turns can contain several independent propositions and are too coarse for
memory-wise intervention. An offline distillation pass rewrites each turn as
atomic third-person statements while preserving names, dates, quantities,
and other attributes used by the questions. The same transformation is
applied to evidence and distractor turns. No answer label is supplied to the
encoding or answering model.

Text-rendered distilled memories and raw turns provide a way to check
whether memory construction loses task-relevant information before KV
encoding. A text-channel result must be distinguished from the accuracy of
injected KV states. A 95\% inclusion threshold cannot be specified reproducibly from the
retained experimental records; accordingly, this value is not asserted as a
selection rule for the reported comparisons. In particular,
the answer-conditioned diagnostic selection described in
Appendix~\ref{app:data} is not treated as a substitute for text-fidelity
validation. No claim of lossless semantic preservation follows from the
act of distilling turns into atomic statements.

For an encoding unit $X$ with $T_X$ tokens, the frozen model stores
\begin{equation}
\begin{aligned}
E_\theta(X;\mathcal C)
&= \left\{(K_\ell,V_\ell)\right\}_{\ell=1}^{L},\\
K_\ell &\in \mathbb R^{T_X\times d_k},\\
V_\ell &\in \mathbb R^{T_X\times d_v}.
\end{aligned}
\label{eq:app_encoding_shapes}
\end{equation}
where $\mathcal C$ is the context visible during cache construction. The
states are produced once and reused across downstream queries whenever the
same encoding unit applies.

\subsection{Matched Visibility Masks}

Let $u(j)$ denote the fact containing token $j$. Independent encoding runs
$E_\theta(m_k)$ separately for each fact. Block-diagonal and causal encoding
instead use the same token sequence, positions, and fact boundaries, with
attention masks
\begin{align}
A^{\mathrm{block}}_{ij}&=
\begin{cases}
0, & u(i)=u(j)\ \text{and}\ j\le i,\\
-\infty, & \text{otherwise},
\end{cases} \\
A^{\mathrm{causal}}_{ij}&=
\begin{cases}
0, & j\le i,\\
-\infty, & j>i.
\end{cases}
\end{align}
The block-diagonal--causal comparison therefore changes cross-fact
visibility while retaining the matched token layout and positions.
Independent bare and independent scaffold encodings process facts separately,
but can differ in contextual scaffolding and other layout details. Their
contrast with causal encoding is not equivalent to the matched
block-diagonal--causal comparison. In particular, no across-model
independent-to-causal gain is interpreted as an isolated visibility effect.

\subsection{Reuse Granularity}

Fact-level units maximize independent addressability. Turn-level units join
atomic statements from one source turn. Session-level units bind relations
across turns. Query-level encoding jointly processes the current candidate
set and is used only as a contextualization reference. If
$N_{\mathrm{unit}}(\mathcal G)$ is the number of addressable units under
granularity $\mathcal G$, broader context generally reduces
$N_{\mathrm{unit}}$ and the reuse factor while increasing the opportunity to
form cross-memory relations. The experiment measures this trade-off; it does
not assume accuracy is monotonic in unit size.

\subsection{Inference-Time Bridge Controls}

Separator, indexed-anchor, and semantic-anchor variants operate on
independently encoded turn-level states after retrieval. They modify the
serving layout or add linking cues but do not recompute stored keys or values.
Session-level encoding is the contextualized reference. The comparison thus
tests whether the evaluated serving cues recover the observed loss; it does
not rule out every possible learned or recomputing bridge.

\section{KV Injection and Allocation Details}
\label{app:injection_allocation}

\subsection{Layer-Wise Cache Construction and Position Correction}

For each layer, retrieved memory banks are concatenated with the query-side
cache:
\begin{equation}
\begin{aligned}
K_\ell^{\mathrm{all}}
&= \operatorname{concat}\big[
K_\ell^q,K_{\ell,1},\ldots,K_{\ell,K}
\big],\\
V_\ell^{\mathrm{all}}
&= \operatorname{concat}\big[
V_\ell^q,V_{\ell,1},\ldots,V_{\ell,K}
\big].
\end{aligned}
\label{eq:app_kv_concat}
\end{equation}
Keys are retained before their serving-time rotary rotation. At injection,
rotary embeddings are applied at the target splice positions, and prompt
positions following the inserted span are shifted accordingly. Values are
reused without modification. The attention mask exposes memory positions to
query tokens while preserving causal decoding. This procedure lets one stored
bank be inserted at different legal positions without retaining a stale
position rotation.

\subsection{Per-Memory Attention Bias}

Let $g(j)$ map injected token $j$ to memory $m_k$. EASE-KV adds one scalar per
memory before the attention softmax:
\begin{equation}
\begin{aligned}
\widetilde z_{tj}^{(\ell)}
&=
\frac{(q_t^{(\ell)})^\top k_j^{(\ell)}}{\sqrt{d_k}}
+b_{g(j)},\\
\widetilde\alpha_{tj}^{(\ell)}
&=
\operatorname{softmax}_{j}
\big(\widetilde z_{tj}^{(\ell)}\big).
\end{aligned}
\label{eq:app_attention_bias}
\end{equation}
Non-memory positions receive zero bias. The same $b_k$ is applied to every
token in memory $m_k$ and, in the main experiments, across all layers. For two
memories, their relative attention odds are multiplied by
$\exp(b_i-b_j)$. The intervention therefore changes competition without
rewriting the stored content. Main controlled experiments use
$\mathbf b\in[-4,4]^K$; this is the evaluated operating range, not a claim
that every optimum lies within it.

\subsection{Scope of Outcome-Aware Allocation Search}

For an evaluated query $q$, an outcome-aware diagnostic may compare a
reference allocation $\mathbf b_q^0$ with alternative vectors in a tested
set $\mathcal B_q$. Such a comparison holds the candidate set, KV states,
model, and answer evaluation fixed while varying allocation. However, it
uses correctness feedback from the very query being assessed. It can
therefore describe instance-level controllability only; it is not a
held-out or deployable allocation policy.

Unlike training-side search for policy distillation, searching allocations
against the current evaluation answer does not constitute held-out policy
performance. We therefore draw no quantitative conclusions from a
per-query oracle-search recovery rate in this paper.

\section{Supervision and Policy Evaluation Details}
\label{app:supervision}

\subsection{Policy Action and Evidence Supervision}

The optional policy maps a query and candidate set to one control per
memory, $\pi_\phi(q,M_q)\rightarrow\mathbf b$, while the answering model
remains frozen. The inspected discrete policy implementation exposes five
slot-level bias actions, $\{-4,-2,0,+2,+4\}$, which lie within the
evaluated $[-4,4]$ control range. This describes the implementation's
action support; it does not establish which exact configuration produced
every reported learned-policy result.

Evidence supervision provides labels $e_k\in\{0,1\}$. It can train a
pointwise or set-interaction selector but does not observe the answer change
caused by a particular $b_k$. Two vectors may agree on evidence membership
and still assign different influence to redundant, conflicting, or bridge
facts.

The evidence-supervised interaction condition predicts relevance while
conditioning on the candidate set. Its selection quality is evaluated
separately from answer accuracy. The downstream supervised-allocation comparison is omitted from the main
allocation table because its original per-question predictions could not be
aligned with a verified run and evaluation denominator. We do not infer
signed-allocation performance from selector metrics alone.

\subsection{Outcome Supervision and Reward Resolution}

For each query, the policy samples $G$ allocation vectors
$\mathbf b^{(g)}\sim\pi_\phi(\cdot\mid q,M_q)$, generates an answer
with the frozen backbone, and receives reward
$r_{q,g}=R(y_{q,g},y_q^*,\mathbf b^{(g)})$. The dependence on the
allocation is important: the inspected training implementation combines
an outcome-correctness component with bias-related penalties, mode-dependent
regularization, and an optional contrast bonus. Schematically, its reward
family has the form
\begin{equation}
    r_{q,g}=r^{\mathrm{correct}}_{q,g}
    -P_{\mathrm{bias}}(\mathbf b^{(g)})
    -P_{\mathrm{mode}}(\mathbf b^{(g)})
    +C_{\mathrm{optional}}(q,\mathbf b^{(g)}),
\end{equation}
where the penalty terms and optional contrast term depend on the selected
configuration. This expression identifies the implemented reward components;
it is not a claim that their exact coefficients or overrides have been
recovered for every reported training run.

The group-relative objective is
\begin{equation}
\begin{aligned}
\mathcal L_{\mathrm{out}}
&= -\mathbb E\Big[
(r_{q,g}-\bar r_q)\\
&\qquad\cdot
\log \pi_\phi(
\mathbf b^{(g)} \mid q,M_q)
\Big],\\
\bar r_q
&= \frac{1}{G}\sum_{g=1}^{G}r_{q,g}.
\end{aligned}
\label{eq:app_outcome_objective}
\end{equation}
This objective is used to probe supervision for memory allocation; it is not
presented as a new RL optimizer.

For binary correctness,
$\operatorname{Var}_g(r_{q,g})=\bar r_q(1-\bar r_q)$. If every allocation in
a group succeeds or every allocation fails, all relative advantages are zero.
We therefore report
\begin{equation}
\rho_{\mathrm{inform}}=
\frac{1}{N}\sum_{q=1}^{N}
\mathbf 1\!\left[\operatorname{Var}_g(r_{q,g})>0\right]
\end{equation}
before aggregating rewards across queries. The reported reward-resolution comparison finds non-zero within-group
contrast for $53.0\%$ of sampled groups under binary correctness and
$73.6\%$ under the evaluated shaped-reward condition. These are
\emph{fractions of informative query groups}, not answer accuracy or
measured policy improvement. The exact run-level sampling count $G$ and
reward-coefficient overrides for this comparison have not been bound to
an executed-run manifest, so no particular launch-script value is assigned
to this result. A separately reported update-level diagnostic associates
$27.9\%$ of shaping pseudo-gradient updates with informative contrast;
its denominator differs from $\rho_{\mathrm{inform}}$ and it should
not be interpreted as a fraction of queries.

\subsection{Initialization and Search--Distill Controls}

Initialization comparisons are intended to change the action prior while
keeping other components comparable. Warm-neutral favors a zero allocation
initially, from-scratch lacks that prior, and warm-legacy uses an earlier
initialization. Different priors can lead to different action distributions
even with similar observed accuracy. Because a complete executed-run
configuration has not been recovered for all policy comparisons, identical
optimization settings across every reported run are not asserted solely
from the presence of shared configuration defaults.

Search--distill replaces on-policy exploration and group-relative credit
assignment with supervised targets obtained from training-side searches.
Failure-only targets use searched training failures, whereas mixed targets
combine searched and reference allocations. The resulting model is evaluated
on the held-out question set without searching its answers per query. The
descriptive test results are interpreted as policy transfer, not as the
success rate of a test-time oracle search.

\subsection{Implementation Defaults and Run Provenance}

The inspected policy code defines a \texttt{SlotInteractionModel} with
implementation defaults of embedding size 512, hidden width 256, two
interaction layers, four attention heads, and dropout 0.1. These values
specify available defaults, not a verified configuration for every
published policy run. An earlier launch note lists $G=8$, while an
inspected rerun script specifies $G=4$. The two are separate experiment
configurations; neither is substituted for the unverified sampling count
of the reported reward-resolution result.

This distinction matters because policy objectives, exploration, and
within-group reward variance may depend on $G$ and on the applied shaping
coefficients. A configuration file or launch command alone does not establish
that a run completed or that its output supplied a particular table entry.

\section{Evaluation and Implementation Protocol}
\label{app:evaluation}

\subsection{Matched Comparison Contract}

The block-diagonal--causal comparison uses identical memory text, token layout, injected
positions, allocation, frozen backbone, prompt, decoding, and evaluator; only
encoding visibility changes. Allocation comparisons reuse the same candidate
set and layer-wise KV states and change only $\mathbf b$ or its predictor.
Supervision comparisons are designed to hold the answering procedure and
policy class comparable while varying learning signals, initialization, or
target construction. The causal interpretation of each matched comparison
applies only to configurations for which those conditions are actually
verified; a shared software default alone is insufficient to establish
run-level equivalence.

Text memory is the delivery reference. Query-level contextualization,
Uniform $\pm2$, and hand-specified recency are diagnostic references:
query-level encoding is not query-independent, and oracle-guided control
uses evidence or temporal information. A hypothetical per-query outcome
search would also be diagnostic, but no unverified quantitative search
result is included. Learned or distilled policies are distinguished by
their use of training-side targets and held-out evaluation without
answer-conditioned test-time search.

\subsection{Models, Decoding, and Judge}

The main frozen answering backbone is Qwen2.5-7B-Instruct~\cite{qwen25}.
Matched encoding comparisons also use Llama 8B and Gemma 12B model families.
Audited experimental paths identify a Llama-3.1-8B family and a Gemma
configuration associated with \texttt{google/gemma-4-12B-it}; a model path,
however, is not an immutable checkpoint revision and does not by itself map
a result row to a completed run. Exact revisions are not claimed where
such mapping is unavailable. The cross-backbone claim concerns the
block-diagonal--causal direction within each family, not absolute
accuracy rankings across architectures.

Answers are generated greedily. The reported main comparisons employ a
Qwen2.5-7B-Instruct semantic-equivalence judge that receives the generated
answer and reference answer, with injection state cleared before judgment.
It is used for evaluation rather than retrieval, encoding, or policy
execution.

Within a matched comparison, the judge and evaluation protocol are held
fixed. Table~\ref{tab:cross_model} reports the block-diagonal--causal
contrast; independently encoded bare or scaffold references are not
silently substituted for block-diagonal rows.

\subsection{Experiment-Specific Dataset and Split Records}

The allocation ablation uses the 200-question evaluation documented in
Table~\ref{tab:app_allocation_counts} (149 single-hop, 51 multi-hop).
A separate evidence-supervised selector experiment has an archived split
of 144 and 56 questions. The separately audited ceiling-gate experiment
records a 750-question training-side pool partitioned into 617 training
and 133 development questions, plus 200 test questions. These numbers
refer to distinct experiment records, not a universal train/dev/test
partition of the entire paper.

\begin{table*}[t]
    \centering
    \small
    \setlength{\tabcolsep}{4.5pt}
    \renewcommand{\arraystretch}{1.08}
    \begin{tabularx}{\textwidth}{@{}>{\raggedright\arraybackslash}p{3.2cm}>{\raggedright\arraybackslash}p{4.0cm}>{\raggedright\arraybackslash}X@{}}
        \toprule
        \rowcolor{black!4}
        \textbf{Evaluation record} & \textbf{Verified counts} & \textbf{Scope and caveat} \\
        \midrule
        Allocation ablation (Table~\ref{tab:allocation_main})
        & Test: 200 (149 single-hop, 51 multi-hop)
        & Four retained methods share the evaluated question IDs; supervised downstream row excluded. \\
        Separate evidence-selector split
        & 144 / 56
        & Experiment-specific training/evaluation split; not the denominator of Table~\ref{tab:allocation_main}. \\
        Answer-conditioned ceiling-gate audit
        & Train: 617; dev: 133; test: 200; selected test subset: 149
        & The subset is chosen using test text-oracle correctness and is diagnostic only. \\
        LongMemEval benchmark
        & Nominal benchmark: 500 questions
        & Run-specific denominators for all derived variants have not been verified. \\
        \bottomrule
    \end{tabularx}
    \caption{Counts grounded in specific evaluation records. Entries with
    different split conventions must not be treated as one shared partition.}
    \label{tab:app_verified_splits}
\end{table*}

The answer-conditioned gate must not be used to describe the composition
of the allocation-ablation test set: both contain the number 149, but one
counts correctness-selected questions and the other counts single-hop
questions. Conversation-disjoint splits can reduce cross-conversation
leakage, yet the mere absence of overlapping conversation identifiers
does not establish that all tuning decisions avoided test information.

\subsection{Injection Calibration}

Before semantic evaluation, native prefill and reconstructed injection are
compared at a bf16 numerical noise floor. Calibration covers layer-wise keys
and values after position correction, downstream query hidden states, answer
logits, and greedy generations. The noise floor is estimated by prefilling
identical text under different batch configurations. These checks validate
the serving path and prevent position or cache assembly errors from being
counted as memory effects; they are not predictive-accuracy evidence.

\subsection{Attention Backend}

Long-context text references use bf16 SDPA attention because eager attention
exhausts memory at approximately 22K tokens. On a symmetric 20-item probe,
SDPA and eager attention differ on three judged items (70\% versus 75\%), with
flips in both directions and fluent generations throughout. We report this
implementation detail to bound cross-backend comparisons. Matched latent
comparisons use a fixed backend within each experiment.

\subsection{Metrics and Statistical Comparisons}

Answer performance is reported as accuracy and percentage-point differences.
Evidence selection uses precision, recall, and PR-AUC. Reward-resolution
analysis additionally reports $\rho_{\mathrm{inform}}$ with query groups
as its denominator; an update-level diagnostic has a different denominator.
The learned-policy comparisons in Table~\ref{tab:learning_main} are
\emph{descriptive}. We do not report hypothesis-test $p$-values or claims
of statistical equivalence, because the complete paired-outcome records
and original test specifications for these comparisons are not established.
Tests performed in separate diagnostic audits are not substituted for
statistics from the reported model comparisons. Diagnostic references and held-out policy results
are not pooled as estimates of the same deployable performance.

\subsection{Software and Compute Reporting Scope}

Separate archived mechanism evaluations record different software settings:
one Qwen mechanism readout records \texttt{transformers} 5.13.1, while a
Gemma mechanism environment records \texttt{transformers} 5.12.1,
\texttt{torch} 2.13.0+cu130, and eager attention. These settings belong to
their corresponding experimental records and are not asserted as a common
environment for every reported result. Immutable model revisions,
complete completed-run parameter overrides, and aggregate GPU hours are
not available for all comparisons; no global compute total is inferred
from individual launch scripts.

\FloatBarrier
\section{Additional Result Decompositions}
\label{app:results}

\begin{table*}[t]
    \centering
    \small
    \setlength{\tabcolsep}{5.5pt}
    \renewcommand{\arraystretch}{1.08}
    \begin{tabular}{@{}lrrrrrr@{}}
        \toprule
        \rowcolor{black!4}
        \textbf{Setting} & \textbf{Text} & \textbf{Latent} & \textbf{Diagnostic}
        & \textbf{Initial gap} & \textbf{Recovered} & \textbf{Residual} \\
        \midrule
        LongMemEval & 51.6 & 31.2 & 38.4 & 20.4 & 7.2 & 13.2 \\
        Full-haystack & 55.5 & 38.6 & 41.5 & 16.9 & 2.9 & 14.0 \\
        Temporal conflict & 77.8 & 56.9 & 76.4 & 20.9 & 19.5 & 1.4 \\
        \bottomrule
    \end{tabular}

    \caption{Decomposition of end-to-end accuracy (percentage points).
    ``Recovered'' is Diagnostic minus Latent; ``Residual'' is Text minus
    Diagnostic.}
    \label{tab:app_gap_decomposition}
\end{table*}

\subsection{Evidence-Selector Audit}

The main full-test selector protocol is conversation-disjoint and removes
source identifiers that could directly reveal the evidence label.
Retraining and leave-one-conversation-out evaluations probe selector
behavior under alternative fitting and evaluation constructions. A separate
ceiling-gated audit uses text-oracle correctness on the test
questions and is discussed in Appendix~\ref{app:data}. Since its
selection protocol conditions on test answers, its results are not included
in the ungated selector comparison table.

\begin{table}[t]
    \centering
    \small
    \setlength{\tabcolsep}{9pt}
    \renewcommand{\arraystretch}{1.08}
    \begin{tabular}{@{}lrrr@{}}
        \toprule
        \rowcolor{black!4}
        \textbf{Protocol} & \textbf{Precision} & \textbf{Recall} & \textbf{PR-AUC} \\
        \midrule
        Full test & 64.50 & 46.02 & 60.62 \\
        Retrain & 69.81 & 40.00 & 60.25 \\
        7-fold LoCoMo & 72.0 & 50.1 & 69.4 \\
        \bottomrule
    \end{tabular}

    \caption{Evidence-selector audit (\%). Selection quality is reported
    separately from downstream answer accuracy.}
    \label{tab:app_selector}
\end{table}

\FloatBarrier

Evidence-selection scores do not uniquely determine signed attention
allocation. Multiple bias vectors can agree on the same evidence labels
while assigning different influence to redundant, conflicting, or bridge
facts. We therefore interpret selector precision, recall, and PR-AUC as
relevance measures, not as a demonstrated improvement in downstream
allocation accuracy. The evidence-selection results alone do not establish that a learned
allocation policy matches an oracle-guided allocation reference.

\subsection{Representation Gap and Allocation Recovery}

The end-to-end settings can be decomposed into the initial text--latent gap,
the amount recovered by a diagnostic allocation, and the residual gap to
text. These values use the same rows as Table~\ref{tab:end_to_end}; they are
shown in Table~\ref{tab:app_gap_decomposition} so that an allocation gain is
not read as evidence of lossless representation.

LongMemEval and full-haystack retain sizeable residual gaps under the tested
allocation controls. Temporal conflict behaves differently: hand-specified
recency recovers most of the gap, showing
that the fixed states can express the desired preference. The learned temporal
policy reaches $70.8\%$, below the $76.4\%$ diagnostic control, again
separating expressivity from learnability.

\subsection{Interpretation of Diagnostic Controls}

The causal--block-diagonal comparison supports an encoding-time visibility
effect under matched layout and positions. The evaluated inference bridges
do not recover the corresponding loss in the reported setup. Evidence-guided
allocation and hand-specified temporal recency show that influence controls
can change answers without modifying stored KV states; because they use
oracle information, they measure diagnostic capacity rather than deployable
performance. The temporal learned-policy result remains below its recency
diagnostic reference. Reward contrast and descriptive held-out policy
performance then address a separate question: whether useful controls can
be identified and generalized from training feedback. The conclusions do not rely on oracle-search rescue rates or
unsupported statistical significance claims.

\end{document}